\documentclass{article}
\usepackage{amsmath}
\usepackage{PRIMEarxiv}
\usepackage[caption=false,font=normalsize,labelfont=sf,textfont=sf]{subfig}
\usepackage{makecell}
\usepackage{bm}

\usepackage{amsmath}

\usepackage[utf8]{inputenc} 
\usepackage[T1]{fontenc}    
\usepackage{hyperref}       
\usepackage{url}            
\usepackage{booktabs}       
\usepackage{amsfonts}       
\usepackage{nicefrac}       
\usepackage{microtype}      
\usepackage{graphicx}
\title{Channel Estimation by Infinite Width Convolutional Networks
\thanks{\textit{{
M. Mallik and G. Villemaud are with INSA Lyon, Inria,  CITI, UR3720, France. (email:mohammed.mallik@insa-lyon.fr, guillaume.villemaud@insa-lyon.fr).This work was supported by a French government grant managed by the Agence Nationale de la Recherche under the France 2030 program, reference “ANR-22-PEFT-0008” - 203831.}}
}}

\author{
{Mohammed Mallik,
and Guillaume Villemaud} \\\\
\small
}
\begin{document}
\maketitle

\begin{abstract}
In wireless communications, estimation of channels in OFDM systems spans frequency and time, which relies on sparse collections of pilot data, posing an ill-posed inverse problem. Moreover, deep learning estimators require large amounts of training data, computational resources, and true channels to produce accurate channel estimates, which are not realistic.
To address this, a convolutional neural tangent kernel (CNTK) is derived from an infinitely wide convolutional network whose training dynamics can be  expressed by a closed-form equation. This CNTK is used to impute the target matrix and estimate the missing channel response using only the known values available at pilot locations. This is a promising solution for channel estimation that does not require a large training set. Numerical results on realistic channel datasets demonstrate that our strategy accurately estimates the channels without a large dataset and significantly outperforms deep learning methods in terms of speed, accuracy, and computational resources.
\end{abstract}


\section{Introduction}
Orthogonal frequency-division multiplexing (OFDM) is a modulation technique that has been extensively adopted in communication systems because it enables while maintaining excellent bandwidth efficiency. It also shows robust capabilities to frequency-selective fading in wireless channels. OFDM divides the available spectrum into a number of overlapping but orthogonal narrowband subchannels, and hence converts a frequency selective channel into a non-frequency selective channel \cite{li2002mimo}. As a result, OFDM has started to be standardized as a key physical-layer technique in wireless LAN standards such as the American IEEE802.11a and the European equivalent HIPERLAN/2 and many commercial systems \cite{9363693}. 

The received signal in a communication channel is typically distorted by the channel's characteristics. Channel effect basically needs to be estimated and compensated at the receiver to recover the transmitted symbols. Generally, the transmitter and receiver both know the positions and values of the pilots in the time-frequency grid, which are symbols used by the receiver to estimate the channel.

Deep Learning (DL) has recently emerged as a powerful tool in communication systems, attracting significant research interest. Various DL-based approaches have been introduced to enhance the performance of traditional algorithms, particularly in tasks such as modulation recognition \cite{o2017introduction}, signal detection \cite{samuel2017deep}, and channel estimation \cite{li2023deep,10683367,li2025crossnet,li2022lightweight}. Notably, several studies have explored the use of deep convolutional neural networks (CNNs) for channel estimation \cite{ahmed20245g,soltani2019deep,gizzini2021cnn,dong2020channel}, where both the input data and estimated channel responses are represented as images, leveraging CNNs' strong feature extraction capabilities. In \cite{ahmed20245g}, a convolutional neural network (CNN) was employed for channel estimation. The authors utilized the 5G New Radio Tapped Delay Line channel model to generate training datasets for this purpose. Their approach involved feeding the real and imaginary components of the Least Square (LS) estimated channels as input to the CNN, which then predicted the full channel response.  
Similarly, in \cite{li2022lightweight}, a hybrid model combining a CNN and a transformer was proposed for channel estimation. In this method, a lightweight CNN was first used to extract relevant features from the pilot data. These extracted features were then passed through a transformer network to refine the channel estimation. Several studies have explored deep learning techniques for channel estimation, considering the problem as an image-to-image translation task. In \cite{dong2020channel}, the authors have considered a conditional generative adversarial network (cGAN) to estimate the channels as images. In that work, the input of the network was the quantized received signal, and the output was the estimated channel. Similarly, Super Resolution Convolutional Neural Networks (SRCNN) have been used for channel estimation in \cite{soltani2019deep, gizzini2021cnn}. In \cite{soltani2019deep}, a two-stage SRCNN model was proposed, taking as input a 2D noisy image representing the channel response at pilot locations and producing the estimated channel as output. In \cite{gizzini2021cnn}, the authors followed a similar approach but introduced a custom pilot arrangement scheme to generate the input pilot images before estimating the channel. Another study \cite{fu2023deep} also leveraged customized pilot placement strategies for deep learning-based channel estimation. In all these works \cite{dong2020channel, gizzini2021cnn, soltani2019deep, li2022lightweight}, the channel response was estimated by training the deep learning CNN models while using the simulated true/groundtruth channels to compute the training loss of the network, which is not available in reality.

Motivated by the above mentioned works, in this work, a lattice type pilot pattern is considered and the time-frequency grid of the channel response is regarded as a three-dimensional (3-D) image. The proposed contribution is to train an infinite width neural network \cite{arora2019exact,williams1996computing} for matrix imputation, which does not need a large dataset or  true channels for channel estimation. The objective is to estimate the unknown values of the channel response while using only the known values at the pilot locations. 

Several recent studies have shown that wider neural networks provide notable advantages in terms of generalization, classification accuracy, and feature learning efficiency \cite{lee2019wide,novak2019neural,nagarajan2019uniform}, but have never been applied to channel estimation. 

Interestingly, artificial neural networks (ANN)s behave like Gaussian processes as the network width tends to infinity \cite{novak2020bayesiandeepconvolutionalnetworks,matthews2018gaussian}, which directly links to kernel methods. The proposed method (noted as Channel-CNTK - Channel Convolutional Neural Tangent Kernel) extends this idea further by using a Convolutional Neural Tangent Kernel (CNTK) \cite{arora2019exact} to impute the sparse time-frequency channel response image.
The contribution of this letter is summarized below:
\begin{itemize}
       
    \item To estimate the unknown values of the channel response from only the channel response at pilot positions, an infinitely wide convolutional neural network for kernel regression is employed. 
    \item Numerical results show that our approach outperforms other deep learning approaches when large training sets or true channels are unavailable for training.
\end{itemize}
The remainder of the letter is organized as follows: Section \ref{sec2} describes the scope and background of the proposed approach, Section \ref{sec3} presents the channel estimation methodology that includes problem formulation, input channel image processing, the infinite neural network channel estimation (Channel-CNTK) method, details of the model, and data sets. Experimental results are presented in Section \ref{results}. The conclusion is given in Section \ref{conclude}.

\section{Background}\label{sec2}
If the time-frequency grid is discretized for channel estimation as a grid of points, the prediction of channel response by interpolating data from existing channel response at pilot locations can be considered as a matrix imputation problem, where the missing matrix element values (the grid points) are estimated from the observed entries (pilot points). 
A number of learning-based methods for imputing missing values rely on supervised algorithms that use datasets with complete observations to find connections between missing and available data. \cite{DBLP:journals/corr/KimLL15b, 10.1145/344779.344972}. Authors of \cite{chu2021unsupervised,arora2019exact, radha} introduced a method for regression, shape completion and reconstruction using incomplete data by infinite width neural networks. Radhakrishnan et al. in \cite{radha}, showed drug response prediction and authors of \cite{arora2019exact} showed regression by infinite neural networks. Interestingly, in \cite{chu2021unsupervised}, authors tackled reconstruction and unsupervised shape completion  using sparse scanned data. In that work, the solution depends on a deep prior inspired by the Neural Tangent Kernel (NTK).
\section{Methodology}\label{sec3}
\subsection{Inverse Problem Formulation}

\begin{figure}[t]
\centering
\includegraphics[scale=0.5]{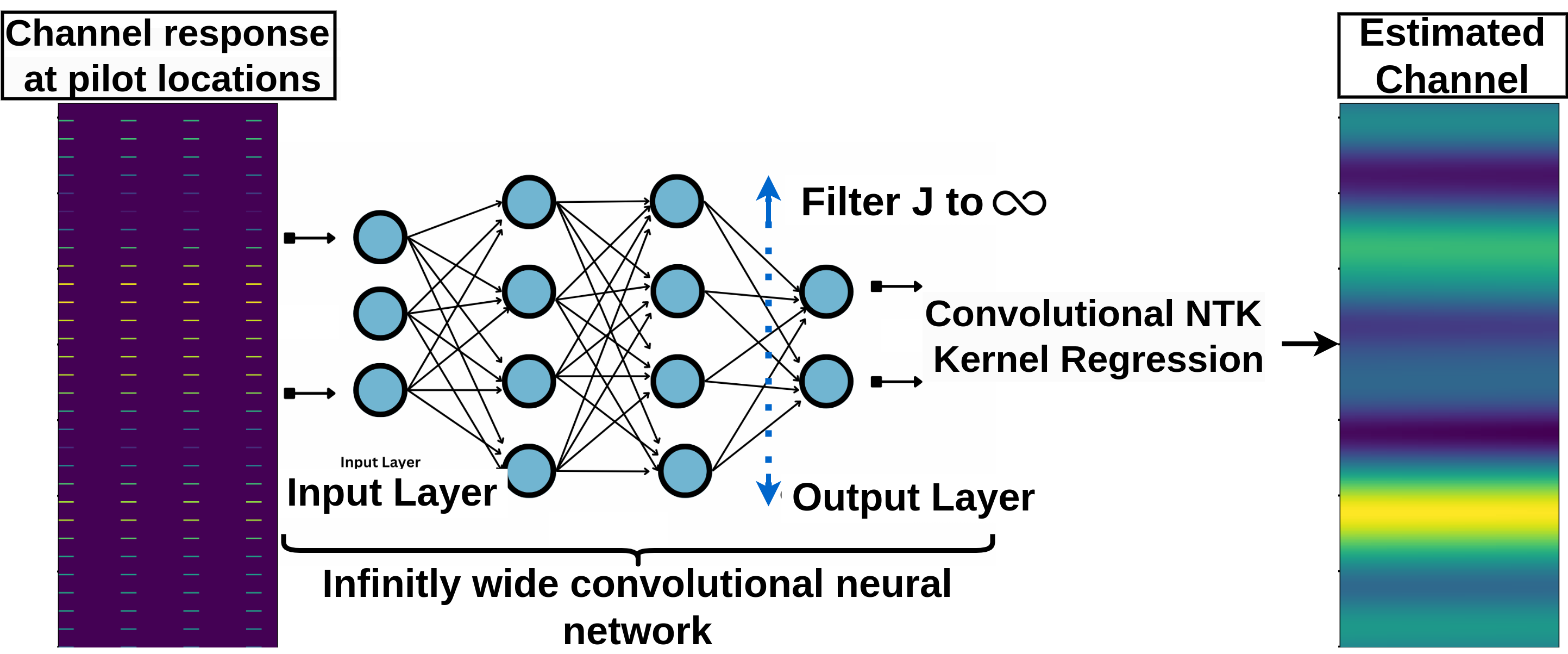}
\caption{An overview of the proposed Channel-CNTK method, which accurately and quickly estimates channel response by exploiting the width limits of neural networks. The main contribution lies in the computation of kernels.}
\label{fig:over}
\end{figure}

In this work, the goal is to estimate the channel time-frequency response $H \in \mathbb{C}^{M \times N}$ within an OFDM frame, where $M$ and $N$ represent the subcarriers and time slots (each time slot contains an OFDM symbol), respectively. The estimation relies solely on sparsely distributed pilot symbols $P$, transmitted at specific subcarriers and time slots. Each channel response $H_{m,n} \in \mathbb{C}^{M \times N}$, located at coordinates $(m, n)$ within the grid, where $m \in \{0, \dots, M - 1\}$ and $n \in \{0, \dots, N - 1\}$. The task of estimating the channel response at $H_{m,n}$ from these limited pilot observations $P$ can be viewed as an inverse problem.
The objective of channel estimation is to learn a function f$_\theta: \mathbb{C}^P \to \mathbb{C}^{M \times N}$ capable of predicting the channel response at all subcarrier-time slot locations where pilots are not available. Here, $\theta$ denotes the parameters of the function $f$, and $P$ represents the number of pilot observations. In this work, $f$ is the infinite width network for kernel regression. A high level overview is shown in Fig. \ref{fig:over}.
\subsection{Image of Channel Response Time–frequency Grid}\label{sec4}
In this study, we focus on a single link—a Single-input, Single-output (SISO) communication link between a pair of Tx and Rx antennas in a fast-moving scenario.
The channel time-frequency response matrix, denoted as 
H (of size $M \times N$) characterizes the link between the transmitter and receiver using complex values. This can be considered as a 3D-image i.e $H \in \mathbb{C}^{C_H \times M \times N}$, where $C_H$, $M$ and $N$ are the channel, height and width of the channel image, respectively. In an OFDM system, the received signal $Y_{m,n}$ after removal of cyclic prefix and discrete Fourier transform performed at the $m$th subcarrier and $n$th time slot can be expressed by Eq. \eqref{eq:1}:
\begin{figure}[!b]
    \centering
    \subfloat[]{%
        \includegraphics[scale=0.20]{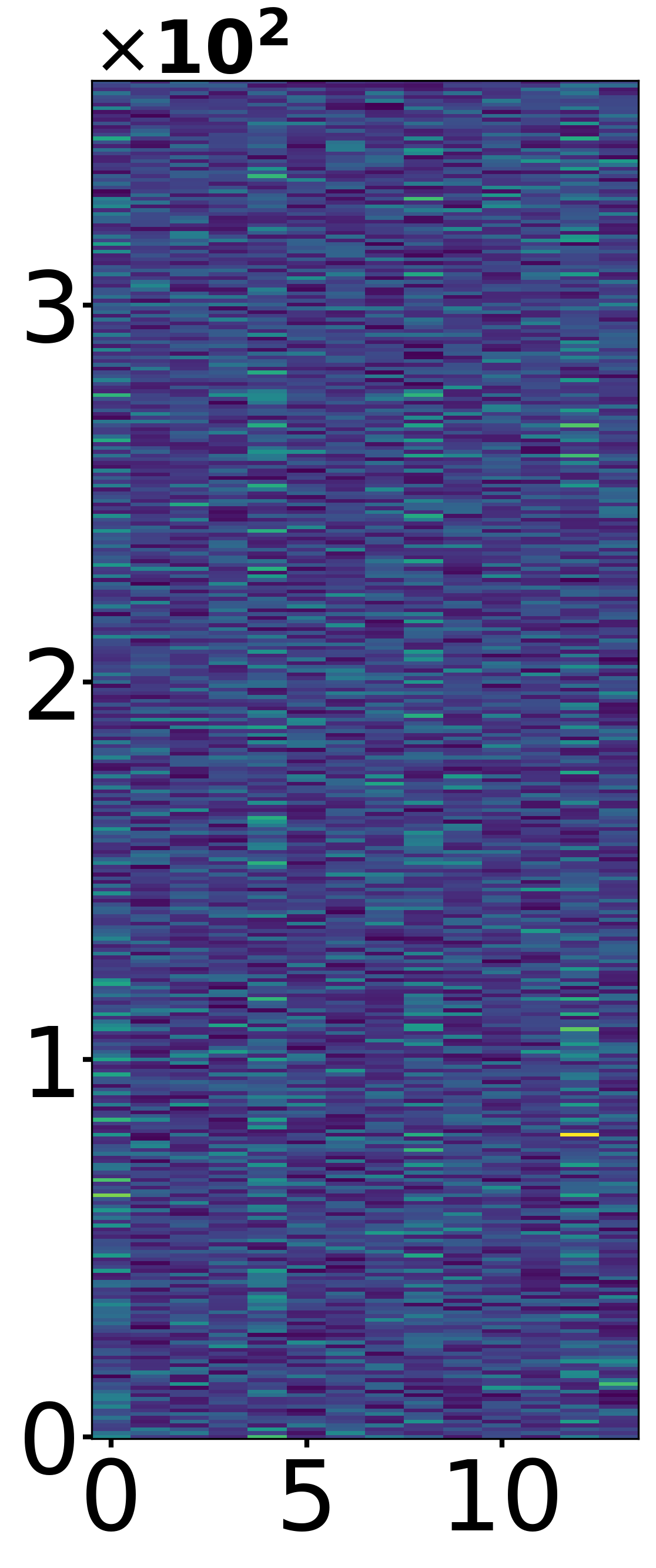}
        \label{fig:received_rxgrid}
    }
    \subfloat[]{%
        \includegraphics[scale=0.20]{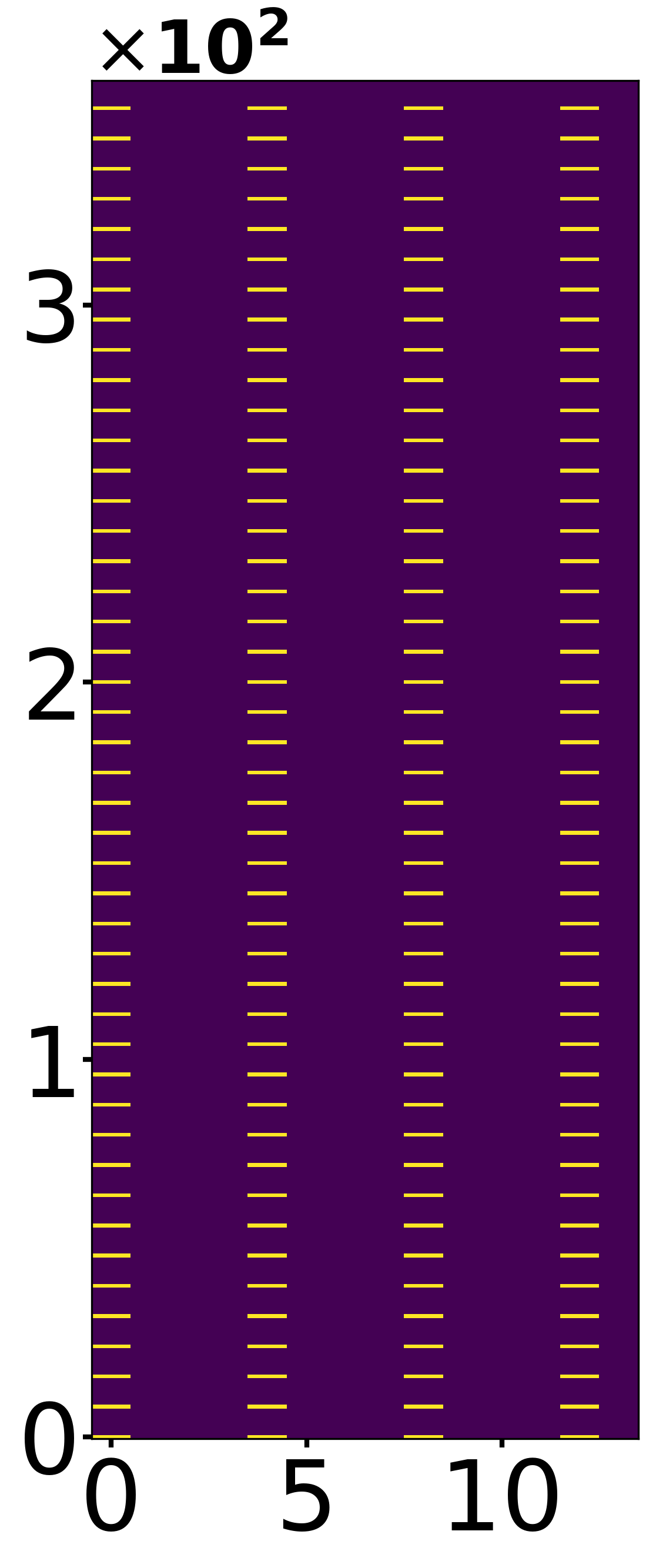}
        \label{fig:transmit_txgrid}
    }
    \hspace{0.2cm}
    \subfloat[]{%
        \includegraphics[scale=0.20]{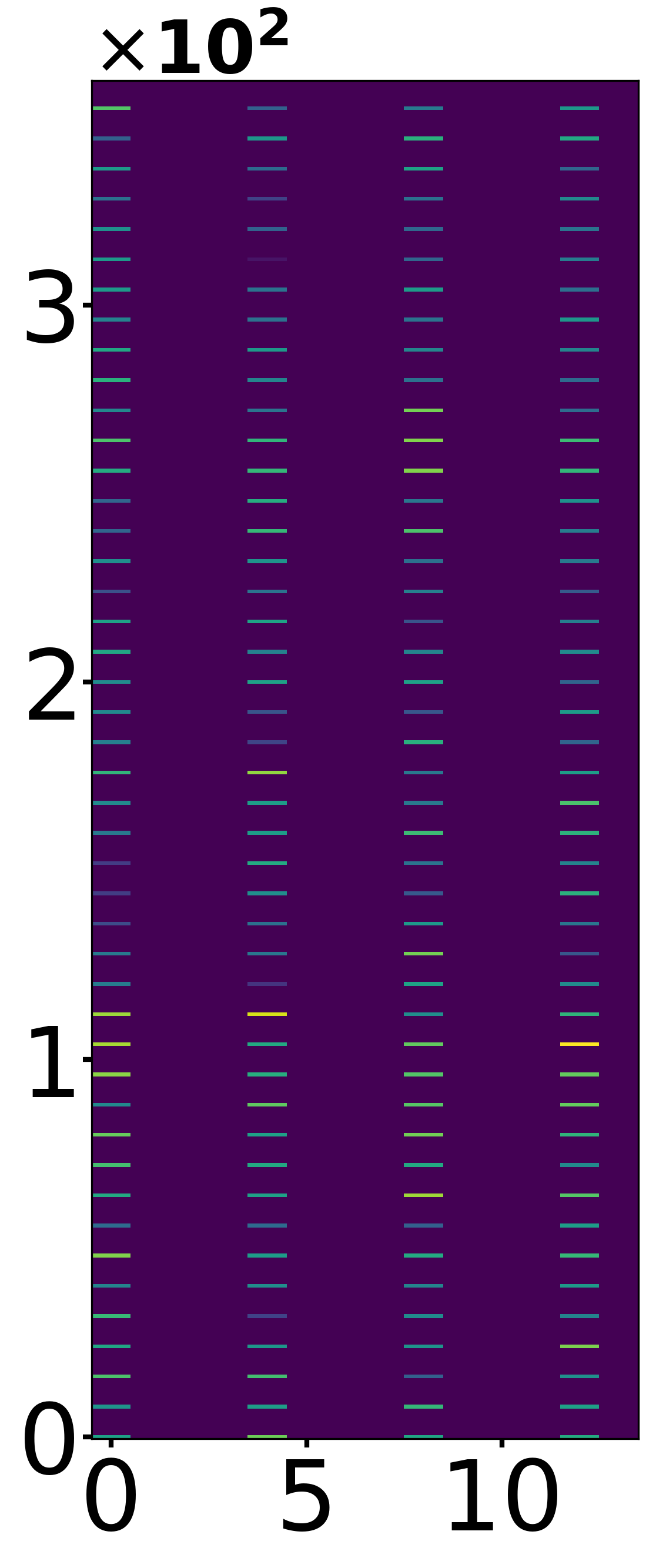}
        \label{fig:h_ls}
    }
    \hspace{0.2cm}
    \subfloat[]{%
        \includegraphics[scale=0.20]{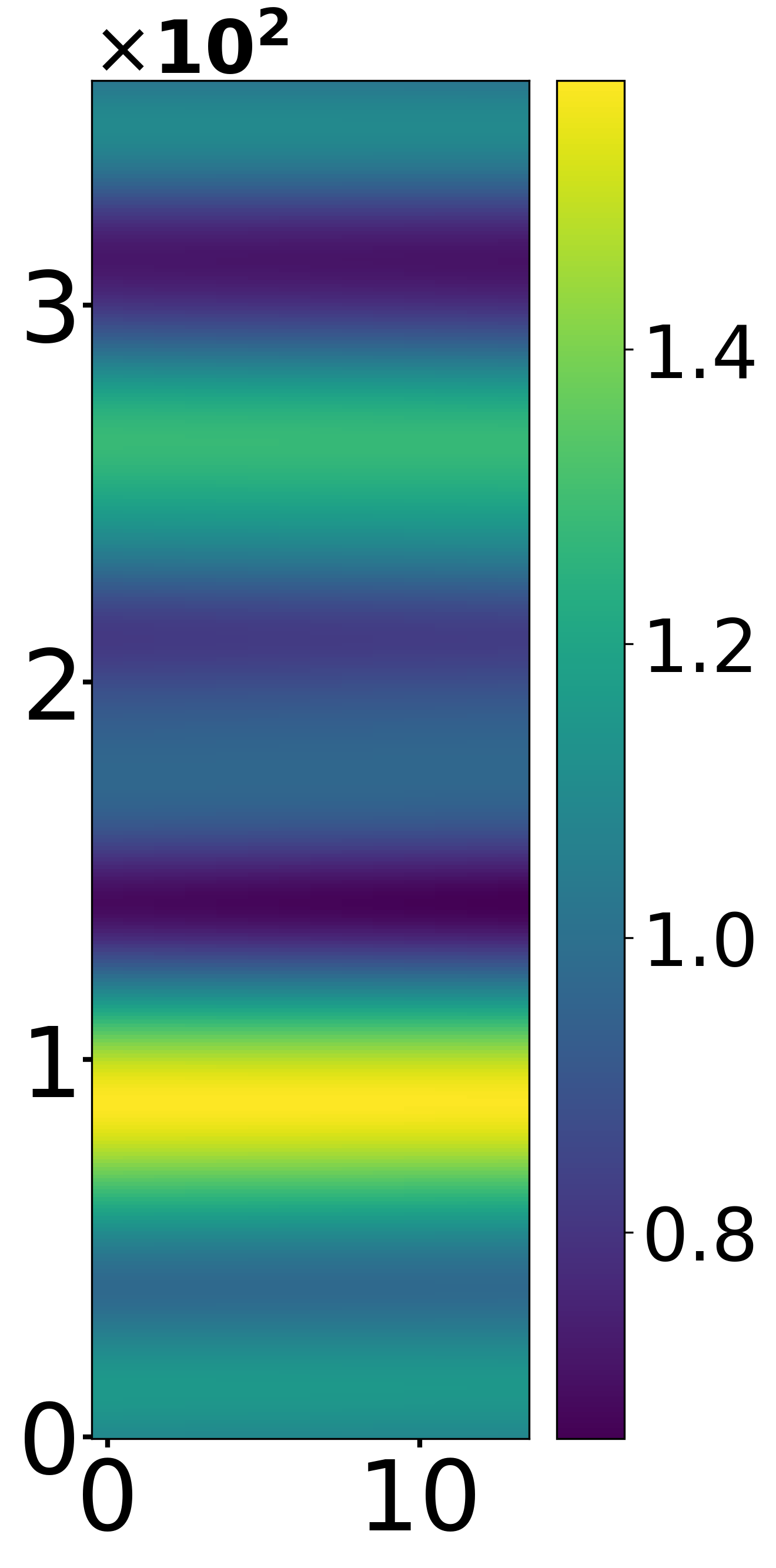}
        \label{fig:perfect_h}
    }
    
    \caption{Example of a sample of channel time-frequency response images. (a) The received signal $Y$, (b) transmitted symbols $X$, (c) channel response at pilot locations $H^p_{LS}$ (the target matrix which we want to impute), (d) perfect channel estimate $H_{perf}$, by MATLAB 5G toolbox \cite{5gToolbox}.}
    \label{fig:cahn_images}
\end{figure}
\begin{equation}\label{eq:1}
Y_{m,n} = X_{m,n} H_{m,n} + Z_{m,n},
\end{equation}

where $Y, X, H, Z \in \mathbb{C}^{M \times N}$ and $X_{m,n}$, $H_{m,n}$,$Z_{m,n}$ represent the known transmitted symbol, channel response, and additive white Gaussian noise (AWGN) with zero mean and variance $\sigma^{2}_w$ at the $m$th subcarrier and the $n$th time slots, respectively. Following the least square (LS) estimation specifically with fading, the estimate of the channel $H^p_{LS}$ at pilot positions can be expressed as :
\begin{equation}\label{eq:2}
    H^p_{LS} = Y / X,
\end{equation}
A visual representation of normalized image for a sample of received signal $Y$, estimated channel response at pilot locations $H^p_{LS}$, (this is the target matrix which we want to impute) and perfect estimation of the channel ${H_{perf}}$ on time-frequency grid with $M$ = 14 time slots and $N$ = 360 subcarriers (based on 5G New Radio (5G-NR) 3GPP standard \cite{5gToolbox}) is shown in Fig. \ref{fig:cahn_images}.

\subsection{Neural Tangent Kernels}

A neural network (NN) is a function $f _{\omega}(x)$  given by: 
\begin{equation}\label{eq:8}
f _{\omega}(x) = \gamma_{\omega_{L}}\phi(\gamma _{\omega_{L-1}}\phi(\gamma _{\omega_{L-2}}\phi(...(\gamma_{\omega_1}(x))...)), 
\end{equation}
where $x$ is the input and $\gamma_{\omega}$ $\in\{1,\cdots,L\}$, are the functions of the layer. To create the layer functions, Neurons (linear function) are followed by a non-linear function known as activation $\phi$, are basic scalar-valued functions that are frequently used. According to the work of Jacot et al. \cite{jacot2018neural}, the behavior of training infinite width networks can be described by a kernel function. Thus, with a kernel known as the neural tangent kernel (NTK) defined below, solving kernel regression is similar to training infinite width networks \cite{jacot2018neural}.\\

\textbf{Definition (Neural Tangent Kernel)}
\textit{Let $f_{\omega}(x): \mathbb{R}^P \rightarrow \mathbb{R}$ denote a neural network with initial parameters $w_0$. The \textbf{neural tangent kernel,} $K: \mathbb{R^d}\times \mathbb{R^d}\rightarrow\mathbb{R}$ is a positive semi-definite function given by:} 
\begin{equation}\label{eq:5}
    K(x,x') = \langle \nabla_{\omega} f_{\omega_{0}}(x),\nabla_{\omega} f_{\omega_{0}}(x')\rangle,
\end{equation}
\textit{where $\nabla_{\omega}f_{\omega_{0}}(x)$ is the gradient and $w_{0} \in \mathbb{R}^P$ denotes the parameters at initialization.}
 
Likewise, for convolutional networks, different number of layers, kernel size, and convolution, skip-connections of neural network are used to construct the CNTK \cite{arora2019exact}. In that case, CNTK kernel for the $L$-th layer is given by:
\begin{equation}\label{eq:6}
    K(x,x') =[\Theta^{L}(A,A')]
\end{equation}
where $\Theta^{L}$ is the convolutional operators defined in \cite{arora2019exact}, 
${A\in \mathbb{C}^{C\times M \times N}}$ is a prior tensor, where $C$ is the number of  channels. In this study, we initialized $A$ using a \textit{prior}, which resembles semi-supervised learning and captures the relationship among the coordinates within the target matrix. In this work, Local Image Prior (LIP) \cite{mallik2024glip,mallik}, which contains the $H^p_{LS}$, is used to initialize $A$.
\subsection{CNN model architecture}

When the network width tends to be infinity, the CNTK is derived using a CNN. Each of the eight convolutional layers in the selected CNN has $ReLU$ activation where 0.05 and 1 are the slopes, respectively. Transposed convolution, upsampling, or trilinear upsampling are employed to change the spatial dimensions of the feature maps, and a stabilization strategy is utilized for numerical stability.

\subsection{Estimating the Channel}
In this case, the sparse channel response $H^p_{LS}$ (see Fig. \ref{fig:cahn_images}) has dimensions $M\times N$. The sub tensors ${H^p_{LS}}_{sub}$ from  $H^p_{LS}$(in the following sections, $M=12, N=14$ is chosen, sub-tensor details are given in section \ref{implementation})  and the CNTK $K \in \mathbb{C}^{M \times N \times M \times N}$. 
Let $\mathbf{X'}$ be the set of available channel response locations and $\mathbf{y}$ the corresponding channel response; $\mathbf{X}$ is the location of an unobserved point where the the prediction of the channel response is made. 
The predicted value $\hat{H}\left(\mathbf{x}\right)$ is given by:
\begin{equation}
    \hat{H}\left(\mathbf{x}\right)=K\left(\mathbf{X},\mathbf{X'}\right)^T.
    K\left(\mathbf{X},\mathbf{X}\right)^{-1}.\mathbf{y} 
    \label{eq:rec}
\end{equation}
where $K\left(\mathbf{X},\mathbf{X'}\right)$ is the CNTK evaluated between the training data and the predicted location. $K\left(\mathbf{X},\mathbf{X}\right)$ and $X$ is the CNTK evaluated using $X$ as the training data.

\section{Numerical Results}\label{results}
In this section, Channel-CNTK approach is evaluated with different pilot density using traditional interpolation techniques and neural network models, namely, ANN and cGAN. 
\subsection{Evaluation Metric}
The normalized mean-squared-error (NMSE) is utilized to calculate the difference between the estimated matrix $\hat{H}$ and
the true channel matrix $H$, which is expressed as:
\begin{equation}
    NMSE = 10 \log_{10} \left\{ \mathbb{E} \left[ \frac{\| H - \hat{H} \|^2}{\| H \|^2} \right] \right\}
\end{equation}
where $||\cdot||$ denotes the matrix norm
calculation and $E$ obtains values of expectation. And $10\log_{10}\{\cdot\}$ is calcualted to obtain NMSE values in decibels.

\subsection{Implementation details and datasets}\label{implementation}
As shown in \cite{ahmed20245g}, the dataset is generated by MATLAB 5G ToolBox\cite{5gToolbox}. For realistic channel modeling and pilot transmission, from the popular MATLAB 5G toolbox, the Tapped Delay Line (TDL) channel \cite{5gToolbox} is employed. MATLAB is used only to create the dataset and was not used elsewhere for the deep learning based channel estimation process. Pilot symbols are placed in every 4th subcarrier and every 2nd OFDM symbol. These parameters are adjusted to increase the density of pilots to 24, 16, and 12 pilots per resource block across of the time-frequency grid to generate the dataset.  The number of resource blocks in our experiments was 30, but it can be 15, 30, etc. The target is to estimate the channel from each sparse $H^p_{LS}$ image. The $H^p_{LS} \in \mathbb{C}^{C_H \times M \times N}$, where $M = 360$, $N=14$, and $C_H=1$. Given the large size of the tensors, we adopted a strategy of subdividing them into smaller sub-tensors ${H^p_{LS}}_{sub}\in \mathbb{C}^{C\times M \times N}$ for the channel-CNTK method. The sparse channel response image $H^p_{LS}$ is split into 30 equal parts as ${H^p_{LS}}_{sub}$ with size $12 \times \ 14 \times 1$ for the imputation process. After the imputation, each sub-tensor was 
stitched together to form the full estimated channel. 
For deep learning models and traditional interpolation methods, they were trained and tested on train and test points of each $H^p_{LS}$. The DNN and cGAN - generator model architecture and parameters are taken and implemented in \cite{rehman2008artificial,abadi2016tensorflow,oskarsson2020probabilistic,virtanen2020scipy} with a GPU backend. 
ANN, cGAN models, and traditional methods were trained on 720 training points of one sample of sparse $H^p_{LS}$. That is approximately only 7 \% of the total pixels available in a $360 \times 14 \times 1$ image. Both ANN and cGAN generator models were trained with 0.001 learning rate, ADAM optimizer and mean squared error (MSE) loss. 
\begin{figure}[!b]
    \centering
    \subfloat[]{%
        \includegraphics[scale=0.18]{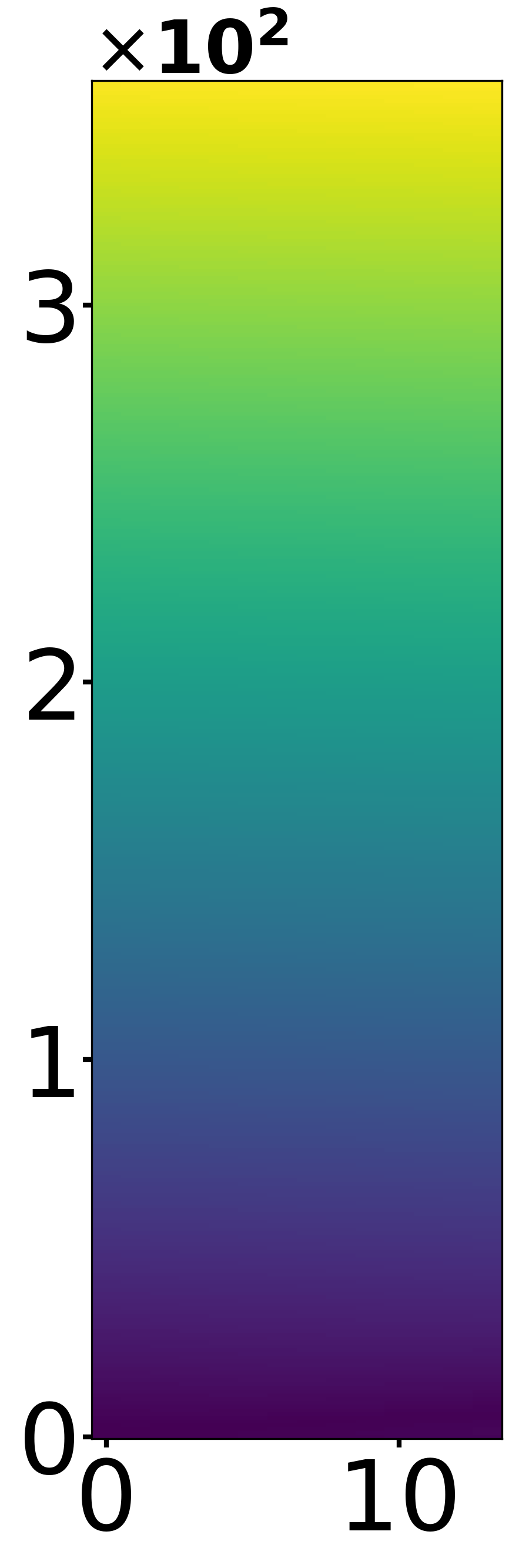}
        \label{fig:dnn}
    }
    \hspace{0.2cm} 
    \subfloat[]{%
        \includegraphics[scale=0.18]{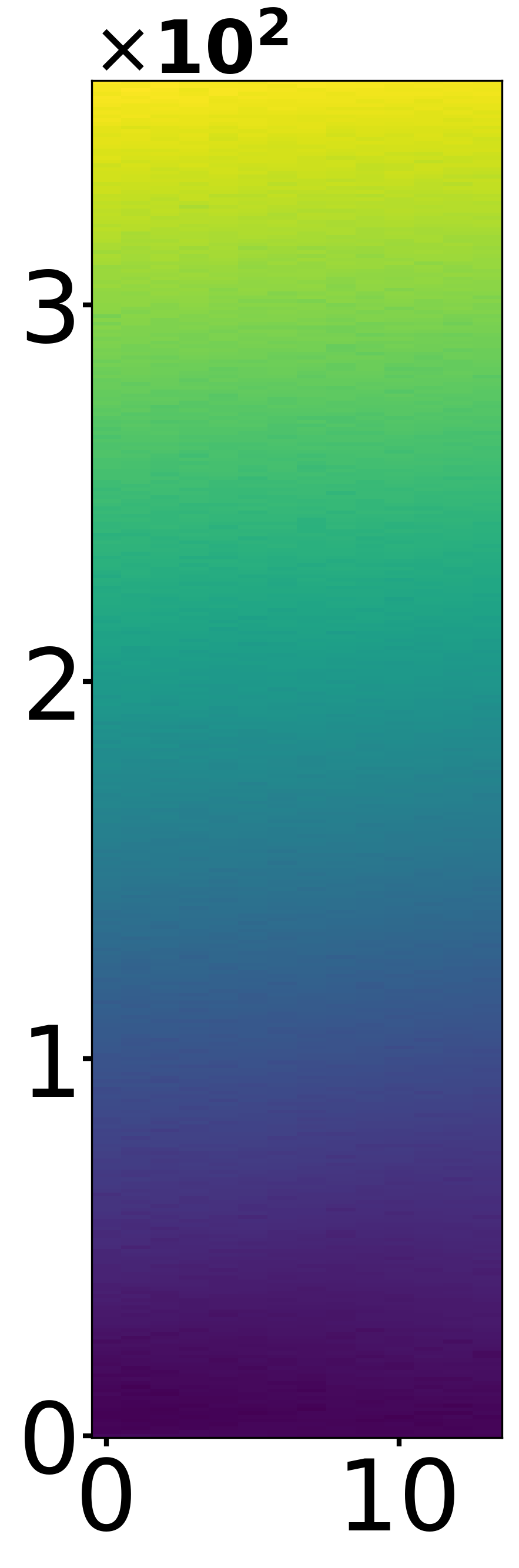}
        \label{fig:cgan}
    }
    \hspace{0.2cm}
    \subfloat[]{%
        \includegraphics[scale=0.18]{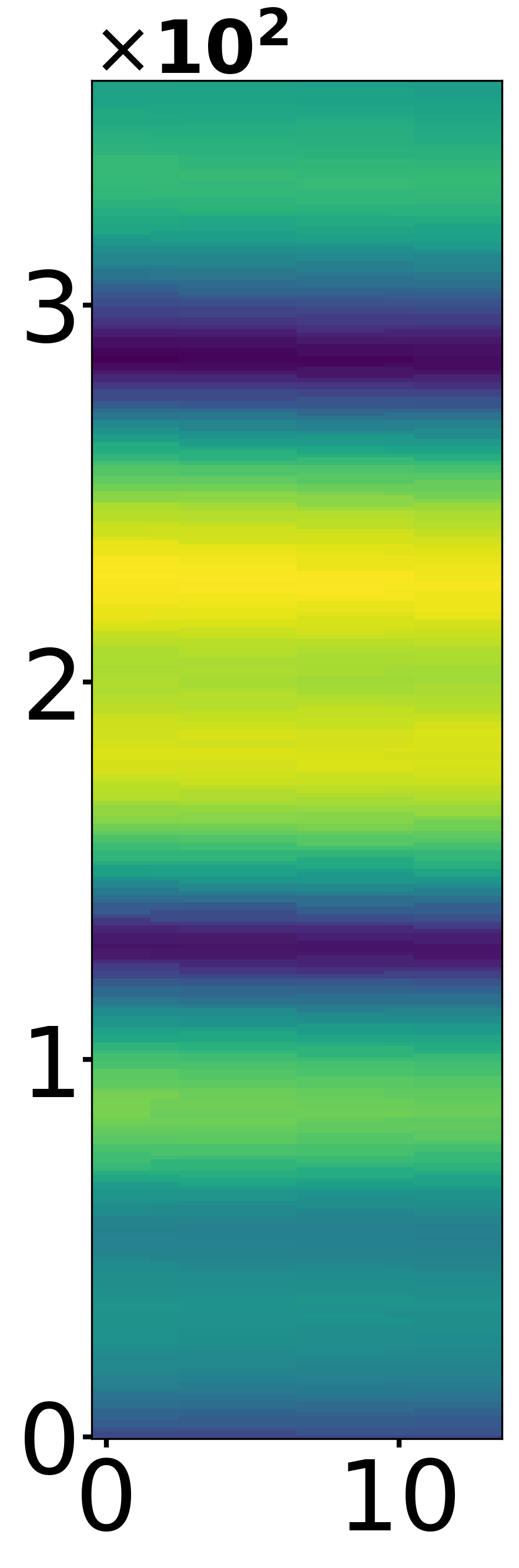}
        \label{fig:nearest}
    }
    \hspace{0.2cm}
    \subfloat[]{%
        \includegraphics[scale=0.18]{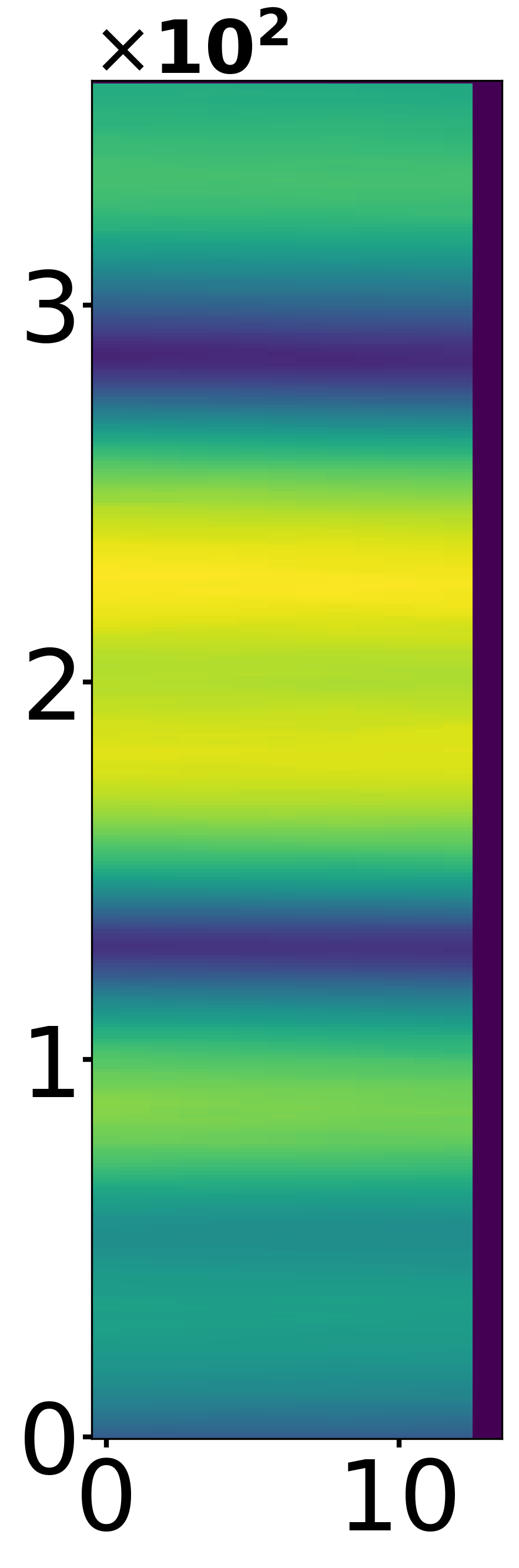}
        \label{fig:linear}
    }
    \hspace{0.2cm}
    \subfloat[]{%
        \includegraphics[scale=0.18]{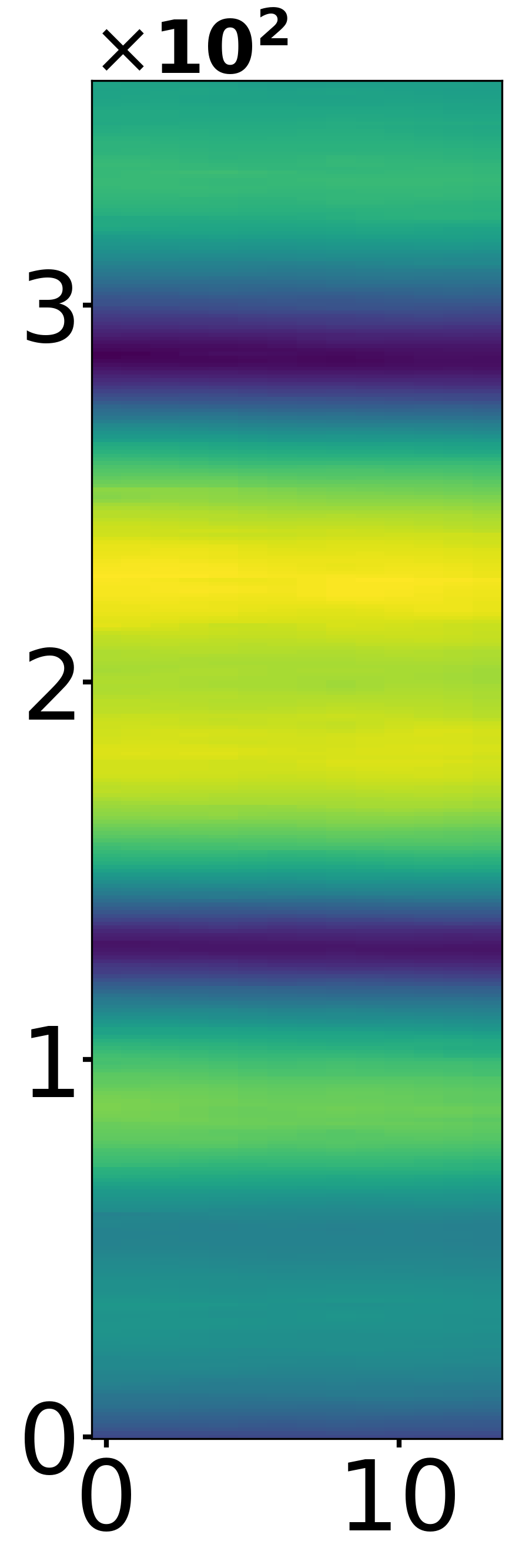}
        \label{fig:ntk}
    }
    \hspace{0.2cm}
    \subfloat[]{%
        \includegraphics[scale=0.18]{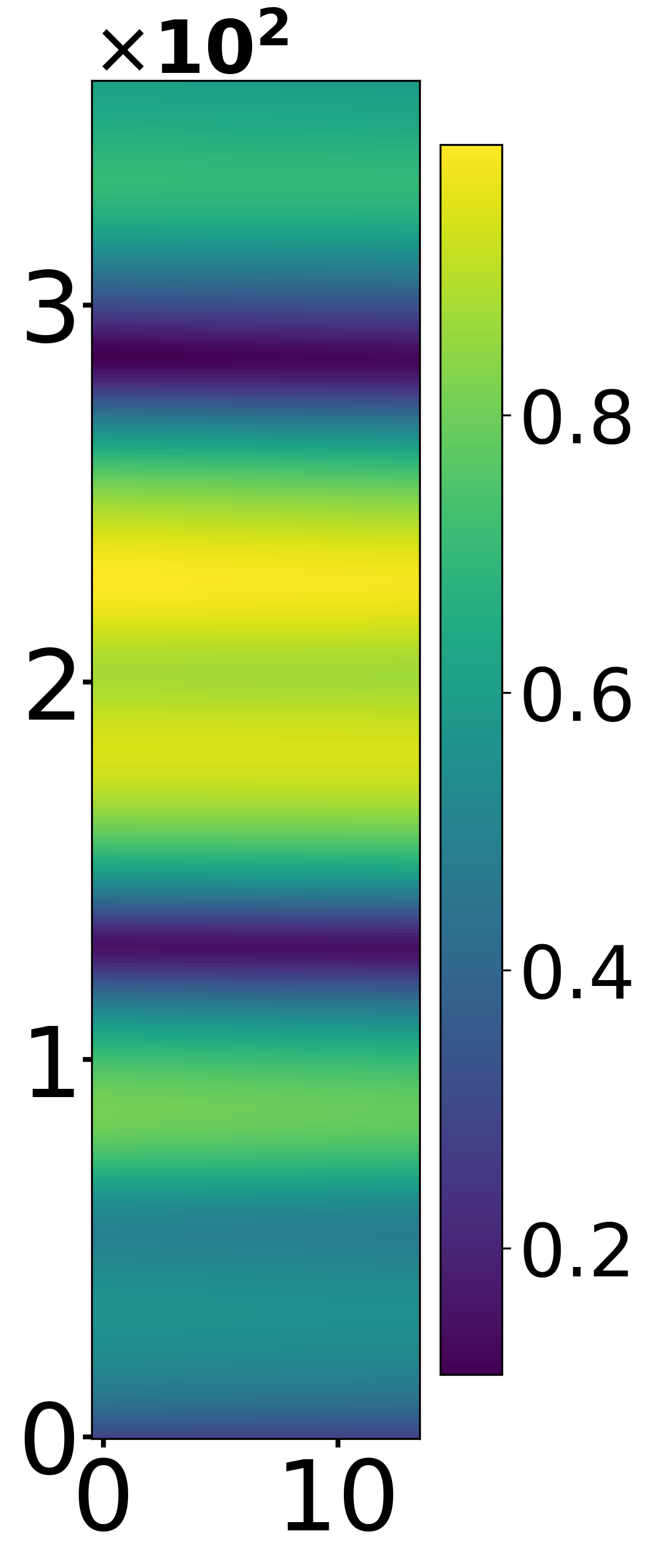}
        \label{fig:perfect_h}
    }
    \caption{Estimated Channels by different methods. (a) and (b) shows the estimation by DNN and cGAN model, (c) and (d) shows the fuzzy and degraded channel estimate from KNN and linear, \textbf{(e)} showcases our Channel-CNTK estimation and (f) is the perfect channel estimation from MATLAB for comparison which was not used for estimation.}
    
    \label{fig:res}
\end{figure}
\subsection{Visual Analysis}
The estimated channels by Channel-CNTK method and comparision with other approaches are shown in pseudo-color images in Figure \ref{fig:res}. The deep models, in Fig. 3(a) (DNN) and 3(b) (cGAN), struggle to generalize and learn the channel features. This is because there are few training samples (24) per resource block that are insufficient to train a DNN or a cGAN with a large number of parameters. As a result, the estimated channels have poor performance, as illustrated in Figure 3(a) and 3(b). Fig. 3(c), KNN estimates the reference (ground-truth) channel response in Fig. 3(f), upon closer inspection (zooming), one can clearly see notable artifacts and fuzzy distortions within the interpolated areas. Linear interpolation in Fig. 3(d), fails to accurately capture the channel characteristics due of its inherent linear assumptions. In contrast, in Fig. \textbf{3(e)} the proposed CNTK-based method shows superior performance, very closely approximating the reference perfect channel response (Fig. 3(f)). The results confirm that Channel-CNTK effectively captures the spatial features of the channel, outperforming deep learning and traditional interpolation methods on a limited number of input data in an image.

\subsection{Quantitative Analysis and Impact of pilot Density}
\begin{figure}[h]
\centering
\includegraphics[scale=0.34]{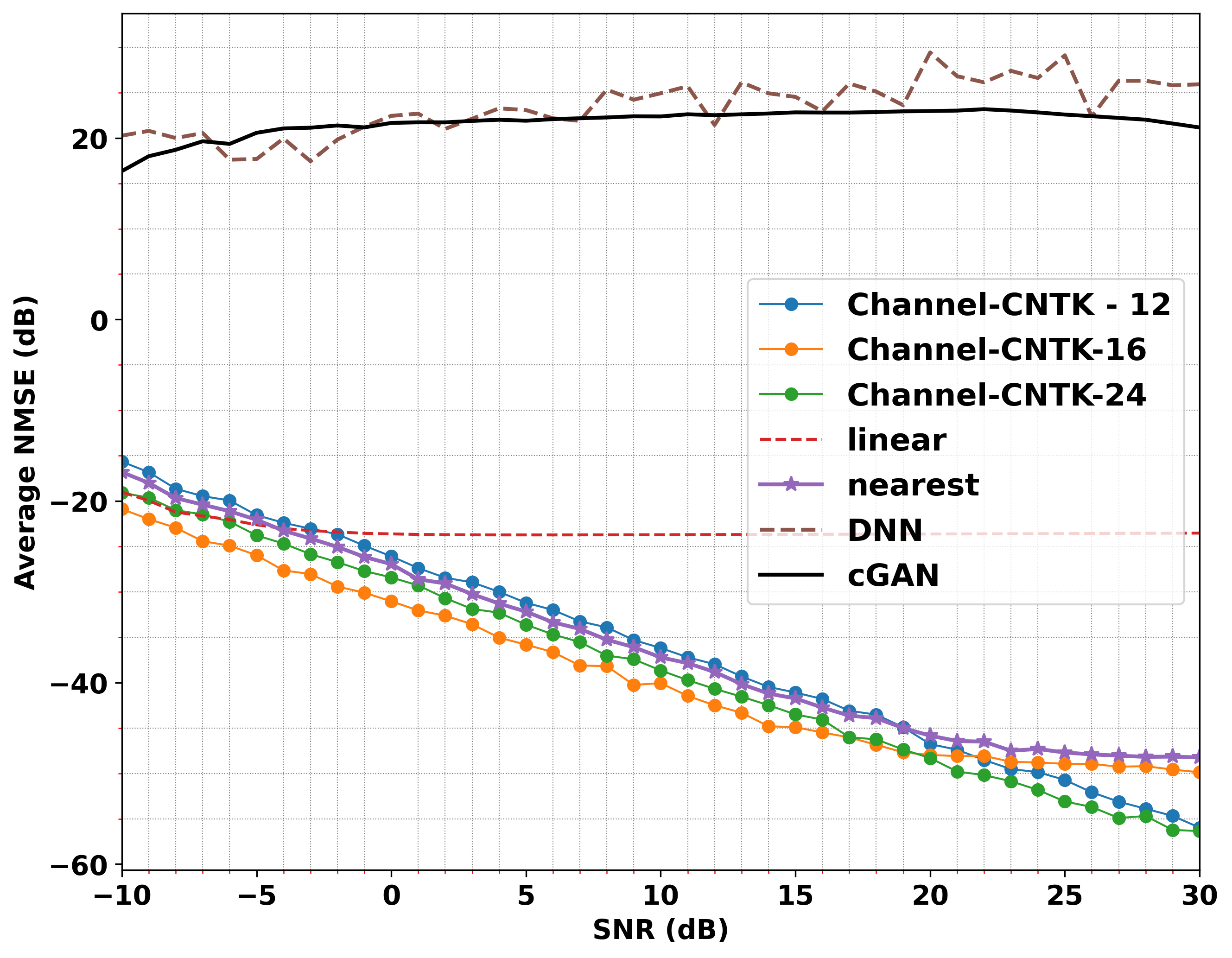}
\caption{Performance of different methods with varying SNRs}
\label{fig:resplot}
\end{figure}

To further validate the outcomes of visual inspection, we compare the impact of various SNR conditions on channel estimateion accuracy. Figure 4 shows NMSE performance of various methods at differnt SNRs  ranging from -10 dB to -30 dB. Our CNTK method outperforms the deep learning models and the traditional interpolation methods in all cases.
As expected, increasing the number of pilot symbols enhances channel estimation accuracy, as shown in Figure \ref{fig:resplot}. Surprisingly, Channel-CNTK yields the best result for all values of SNR even when there are only 24 pilots per resource block. Remarkbly, CNTK demonstrates robust performance at low SNRs than deep learning and conventional interpolation models. This can be attributed to the ability of the infinite-width network to extract feature maps from the structure of the data and provide accurate estimations of the CNTK, therefore mitigating the influence of noise on kernel regression. Conventional
methods such as KNN deteriorates in comparison to CNTK
for all values of SNR, proving the reliability of our
CNTK approach.

\subsection{Comparative analysis of time efficiency of the algorithms}
\vspace{10pt}  

We examined the efficiency of the CHannel-CNTK method in terms of memory consumption and the time required for inference. Our analysis shows that the cGAN-based approach, DNN-based approach required 46 minutes and approximately 46 seconds on one image to train, respectively, on a machine with 32GB RAM and 32GB GPU memory (Intel core i7, NVIDIA ADA). cGAN model iterated over 500 epochs, while DNN trained for 1000 epochs. Both models required higher memory, specifically for DNN 320 MB VRAM and and 10GB VRAM for cGAN for training. In contrast, as shown in Table \ref{tab2}, to train and impute the channel matrix, Channel-CNTK method took only $1,48.10^{-4}$ seconds.

\begin{table}[h]
\begin{center}
\caption{Computational Complexity}
\label{tab2}
\begin{tabular}{|c | c | c | c |}
\hline
Method & Memory Usage &\makecell{Training/\\image} & \makecell{Inference/\\image}\\
\hline
DNN & \makecell{1.2 GB RAM + \\320 MB VRAM} & 40s & 1s\\
\hline
cGAN & \makecell{10 GB RAM \\ + 10 GB VRAM} & 46 min & 1s\\
\hline
Linear& 1,2 GB RAM & -&$1,85.10^{-2}$s\\
\hline
KNN & 1,2 GB RAM & -&$1,846.10^{-2}$s\\
\hline 
\textbf{Channel-CNTK} & \textbf{1,2 GB RAM} &-& \bm{$1,48.10^{-4}$}s \\
\hline
\end{tabular}
\end{center}
\end{table}
\section{Conclusion}\label{conclude}
We have presented a channel estimation framework using an infinitely wide convolutional neural network for CNTK. In contrast to alternative channel estimate methods, this method greatly reduces the time required and computational resources (memory, machine configuration, etc.). Moreover, a large dataset of any reference true/groundtruth channels generated from a simulator is not required for Channel-CNTK as required for training deep learning models. From only few sparsely located pilots in an image, the proposed CNTK based method can estimate the channel response fast with accuracy. In conclusion, the Channel-CNTK is solution for channel estimation, offering a balance between efficiency, computational resources and accuracy when no large dataset or groundtruth is available. Channel-CNTK is a solution which can be extended to uplink, downlink, or massive MIMO channel estimation, with further refinements and validations planned to enhance its robustness and applicability in future.

 
%

\bibliographystyle{IEEEtran}
\bibliography{biblio.bib}

\end{document}